\documentclass[runningheads]{llncs}
\usepackage[T1]{fontenc}

\usepackage{graphicx}
\usepackage{amsmath}
\usepackage{cleveref}

\begin{document}

\title{Orthogonal Discrepancy Kernels for Learning with Partial Physics}

\author{Swapnil Manna\inst{1}\orcidID{0009-0007-7856-9714} \and
Timothy J. Rogers\inst{2,3}\orcidID{0000-0002-3433-3247} \and
Lawrence Bull\inst{3}\orcidID{0000-0002-0225-5010}}
\authorrunning{S.\ Manna, T.J.\ Rogers \& L.A.\ Bull.}

\institute{IISER Pune, India\\
\and
University of Sheffield, UK\\
\and
University of Glasgow, UK \\
}

\maketitle              
\begin{abstract}
We introduce a semi-parametric framework for nonlinear system identification, which decouples discrepancy functions from physics-based components. %
Orthogonal Gaussian process regression balances sparse parameter selection (the white box) with discrepancy learning (the black box) to produce interpretable models from incomplete physics.

\end{abstract}

\section{Introduction}
Across science and engineering, dynamical systems (phenomena governed by differential equations) are ubiquitous. %
Understanding these systems remains an important challenge for researchers and practitioners alike. %
System identification aims to represent such systems given observed data, which is a difficult task, as observations through time depend on previous inputs and states, which are often partially observed. %
One subset of system identification tools also considers the \textit{inverse problem}, to uncover the physical mechanisms of the underlying process as well as a predictive model. %
In other words, the ``true'' system is learnt, in the sense that model parameters are identified which correspond to some physics-derived equations. %
There are significant benefits of this approach, especially regarding extrapolation beyond the bounds of training data. %
Of course, this viewpoint is predicated on the belief that there exists an underlying process which can be mathematically described. %
In this work, we restrict the problem to learning sets of ordinary differential equations (ODEs) for ease of exposition, noting that very similar ideas can be applied in the case of distributed systems with partial differential equations (PDEs), e.g.\ see \cite{MESSENGER2021110525}.

Various schemes can be used to achieve this learning task, often cast as a combined model selection and inverse problem. %
The challenge is to jointly determine: (i) the model structure (i.e.\ which terms to include in the ODE) and (ii) the coefficients of those terms.  %
The original approach was to treat this as a greedy joint optimisation problem, which is typically infeasible (for most realistic systems) since it requires many simulations via some numerical integration scheme, e.g. Runge-Kutta. %

\textit{Sparse approaches} resolve this by initially estimating the states of the system, usually via numerical differentiation/integration of an observed variable. %
Then, one can construct a dictionary of candidate terms, which are basis expansions of the states, e.g. polynomial transformations. %
This re-formulates system identification as a sparse linear regression (far more efficient than greedy optimisation). %
Assumptions must be made: (i) terms with very small coefficients are unimportant and can be removed, (ii) the dictionary of basis covers all the possible latent dynamics. %
Under these assumptions, coefficient ``switching'' is realised by inducing sparsity on the parameter vector, either with a penalised loss function \cite{CORTIELLA2021113620}, sequential pruning of small values \cite{brunton2016discovering} or via sparsity-inducing Bayesian priors \cite{fuentes2021equation},\cite{nayek2021spike}. %
This mechanism for system identification is known widely as SINDy \cite{brunton2016discovering}.

\subsection{Contributions}
In this work, we address shortcomings of SINDy.  
Namely, the potential for bias when the data are corrupted by structured error, typically resulting from "partial physics", where necessary basis functions are missing from the dictionary (i.e. the model discrepancy problem).

Our proposed solution increases the expressivity of the sparse regression, fusing a physics-based dictionary with a flexible machine learner. %
This hybrid construction is usually considered to be ``physics-informed'' or a ``grey-box'' \cite{karniadakis2021physics},\cite{cross2024spectrum}. %
Specifically, a Gaussian process (GP) model represents dynamics that the dictionary cannot, where the GP is constrained to be orthogonal to the dictionary, to favour the physics-inspired bases over the black-box (where possible). We also highlight the advantages of an Orthogonal kernel GP  compared to a standard squared-exponential kernel GP.

\section{Grey-box SINDy}
Consider a first-order dynamical system,
\begin{equation}
\dot{\mathbf{x}} = f(\mathbf{x}),\label{eq:first order dynamical eq}
\end{equation}
where the dynamics $f$ are unknown, but we have prior knowledge of the potential structure.
One might decompose these dynamics into a system matrix $\mathbf{A}$ with an additive discrepancy $\tilde{f}$, 
\begin{equation}
\dot{\mathbf{x}} = \mathbf{A} \mathbf{x} + \tilde{f}(\mathbf{x}) \label{eq:greybox}
\end{equation}
Constructing $\mathbf{A}$ parametrically and $\tilde{f}$ nonparametrically, \cref{eq:greybox} is cast as a \textit{grey-box} model, combining known physics (white-box) with data-driven corrections (black-box). %
This is convenient when first principles suggest an approximate structure for $\mathbf{A}$, while $\tilde{f}$ captures unexpected nonlinearities or missing physics.

We extend \cref{eq:greybox} by replacing $\mathbf{A}\mathbf{x}$ with sparse regression over a dictionary of candidate basis functions. %
Where ${\mathbf{\Phi}}(\mathbf{x}) = [\phi_1(\mathbf{x}), \ldots, \phi_K(\mathbf{x})]^\top$ denotes an overcomplete library of $K$ physics-informed basis,
\begin{equation}
\dot{\mathbf{x}} = {\mathbf{\Phi}}(\mathbf{x})^\top {\boldsymbol{\theta}} + \tilde{f}(\mathbf{x})
\label{eq:sparse_grey}
\end{equation}
sparse coefficients ${\boldsymbol{\theta}}$ are inferred (SINDy \cite{brunton2016discovering}) using an appropriate prior (e.g.\ Laplace) to identify relevant physics, while $\tilde{f}$ is approximated with a Gaussian Process \cite{gpml} (GP) to represent any missing structure from $\mathbf{\Phi}$.

\subsection{Greedy machine learners}

Additive grey-box models present a balancing act between the parametrised component and the machine-learned component. %
Typically, when inferred jointly, the black-box attempts to represent the complete signal, reducing the white-box contributions to zero. %
This imbalance is widely acknowledged for hybrid learners, across GPs, neural networks, and other black boxes. %
With a sparse white-box (\ref{eq:sparse_grey}), this problem is exacerbated, since shrinkage on $\theta$ hands more signal to the (greedy) black box. %

\subsection{Orthogonal kernels}

To protect against greedy machine learners, we constrain the GP to the orthogonal complement of the physics-informed dictionary, to satisfy, %
\begin{equation}
\int_{\mathcal{X}} \mathbf{\Phi}(\mathbf{x}) \tilde{f}(\mathbf{x}) \, d\mathbf{x} = \mathbf{0} \label{eq:constraint}
\end{equation}
This constraint prevents  $\tilde f$ from learning any structure that could be approximated by the dictionary. %
To impose \cref{eq:constraint}, we first place a GP prior on the discrepancy, following \cite{koh}, %
\begin{equation}
\tilde{f} \sim \mathrm{GP}(0, k_\perp)
\end{equation}
An orthogonal discrepancy kernel function $k_\perp$ can be described following \cite{plumlee}, %
\begin{equation}
k_\perp(\mathbf{x}, \mathbf{x}') = k(\mathbf{x}, \mathbf{x}') - \mathbf{\Phi}(\mathbf{x})^\top \mathbf{H}^{-1} \mathbf{\Phi}(\mathbf{x}'),
\end{equation}
where $k(\mathbf{x}, \mathbf{x}')$ is any standard covariance function %
and ${\mathbf{H} = \int_{\mathcal{X}} \mathbf{\Phi}(\mathbf{x}) \mathbf{\Phi}(\mathbf{x})^\top \, d\mathbf{x}}$ is the Gram matrix of basis functions. %
Here, we use the squared-exponential kernel, with dimension-specific length scales for automatic relevance determination \cite{gpml} and Monte Carlo integration to approximate $\mathbf{H}$.

\section{Numerical example}

\begin{figure}
    \centering
    \includegraphics[width=.52\linewidth]{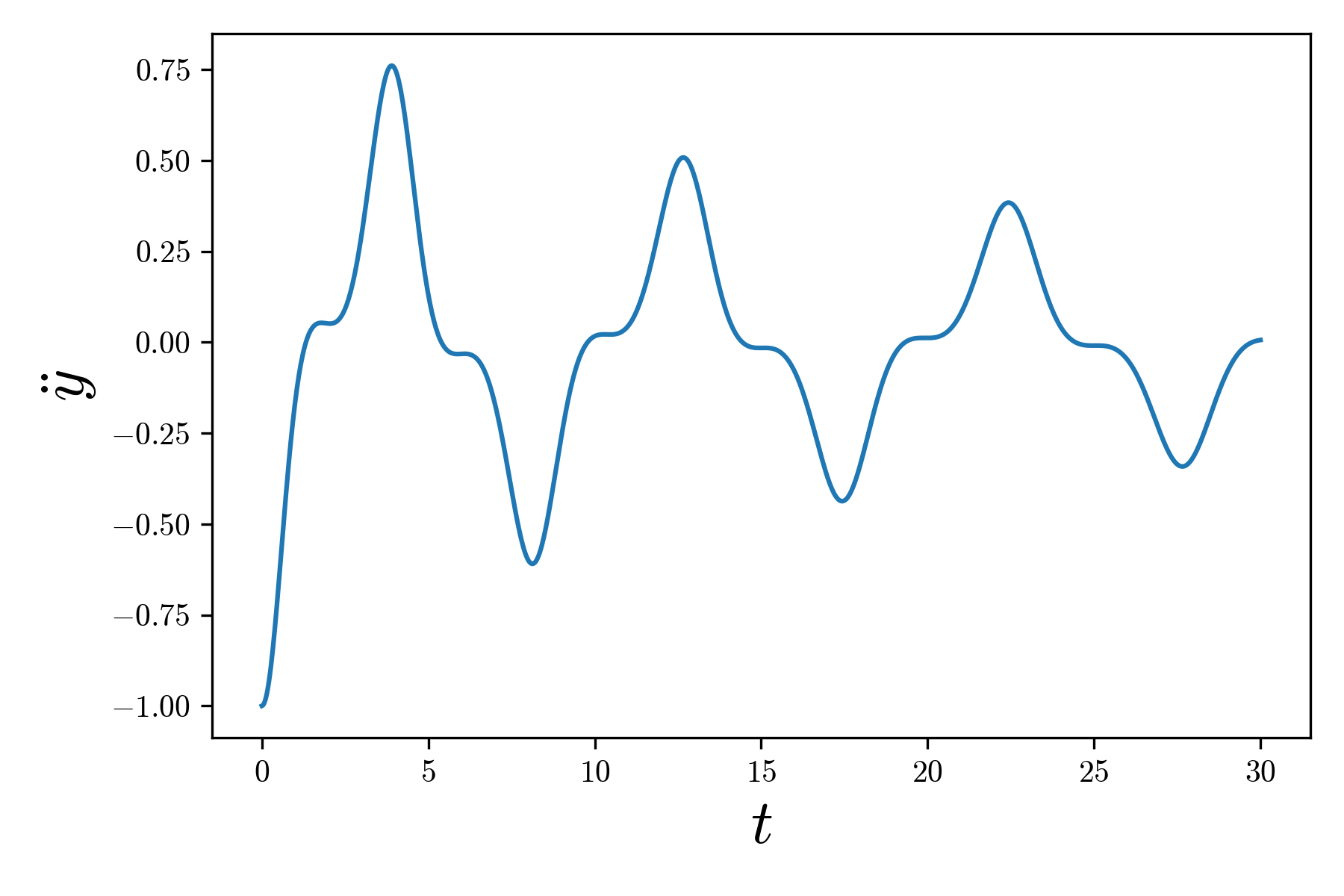}
    \caption{The acceleration response $\ddot y$ of a numerically simulated system with cubic nonlinearity. Showing oscillations through time $t$.}
    \label{fig:cubic-trajectory}
\end{figure}

\noindent For demonstration, consider an (unforced) oscillator with cubic nonlinearity,
\begin{align}
\ddot{y} + c \dot{y}^3 + k y^3 = 0 \nonumber \\ 
\ddot{y} = - c \dot{y}^3 - k y^3 \label{eq:mkc}
\end{align}

\noindent where $c$ and $k$ are damping and stiffness coefficients respectively. %
To present \cref{eq:mkc} in the structure of \cref{eq:first order dynamical eq} one uses state variables $\mathbf x_t = [y(t), \; \dot y(t)]^\top$,

\begin{equation}
\begin{bmatrix} \dot{x_1} \\ \ddot{x_2} \end{bmatrix} = 
\begin{bmatrix} \dot{y} \\  -c\dot{y} - ky^3 \end{bmatrix}.
\end{equation}

\noindent this system of ODEs can be treated as an initial value problem and solved numerically. %
Here, we simulate the system using adaptive Runga Kutta, then treat the acceleration trajectory as `observed' data, shown in \Cref{fig:cubic-trajectory}.

\noindent We apply both, (i) the standard squared-exponential kernel (Non-Orthogonal) GP with SINDy and (ii) the Orthogonal kernel GP with SINDy to our setup and present the results for both complete and incomplete dictionaries.

\paragraph{System Identification using Complete Dictionary: }

Our dictionary of candidate basis functions consists of,

\begin{equation}
\mathbf{\Phi}(\mathbf{x}) = [{y} , {\dot y} , {y}^2 , {\dot y}^2 , {y}^3 , {\dot y}^3]^\top
\end{equation}
\noindent Note how rows of $\mathbf{\Phi}$ are transformations of the state trajectories $\mathbf x_t = \{y(t), \; \dot y(t)\}$.

\begin{figure}
\centering
\includegraphics[width=\textwidth]{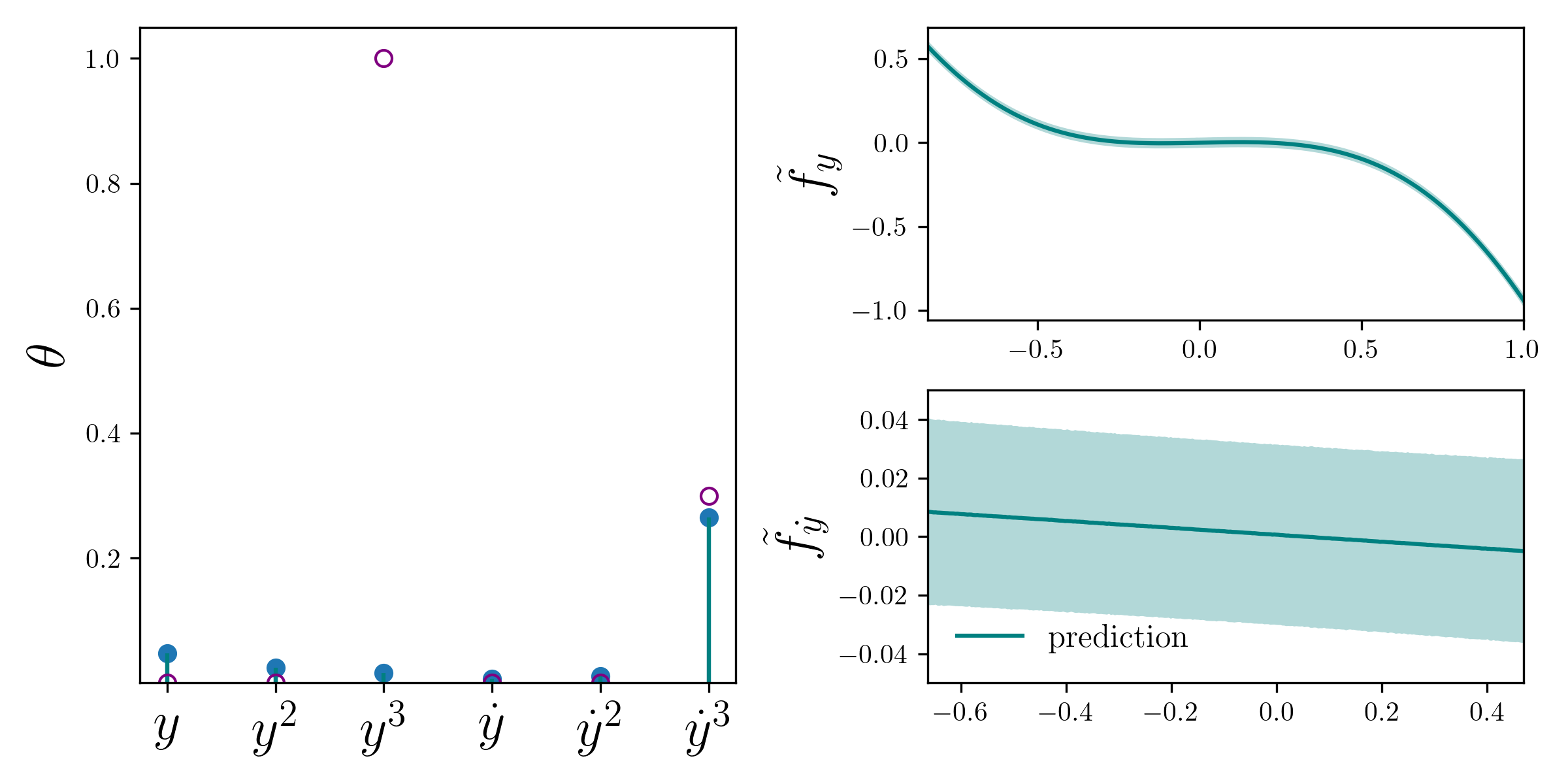}
\caption{System Identification using Non-Orthogonal GP Grey Box Modeling for Complete Dictionary: (left panel) parameters identified via sparse regression using the white box; (right panel) the black-box discrepancy model, which greedily learns the cubic spring basis function (note the different scales of the response).} \label{fig:nonorth}
\end{figure}

\begin{figure}
\includegraphics[width=\textwidth]{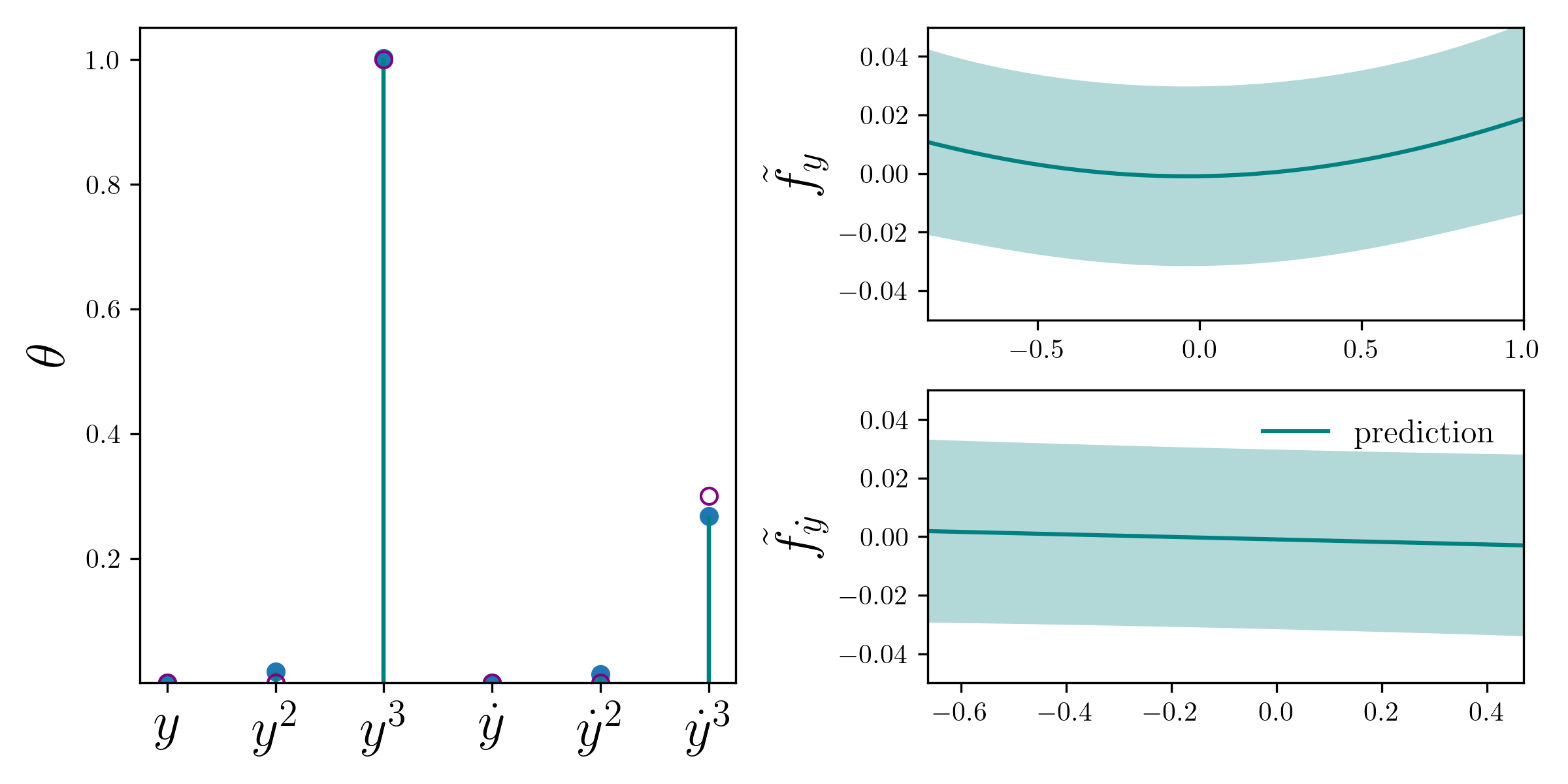}
\caption{System Identification using Orthogonal GP Grey Box Modeling for Complete Dictionary: (left panel) parameters identified via sparse regression using the white box; (right panel) the black-box discrepancy model (note the low scale of the response)} \label{fig:ortho}
\end{figure}

\noindent \Cref{fig:ortho} and \Cref{fig:nonorth} represent the orthogonal and non-orthogonal Gaussian Process (GP) models, respectively.

\noindent From \Cref{tab:placeholder}, we observe that both the methods achieve comparable predictive accuracy. As seen in \cref{fig:ortho}, the orthogonal kernel GP yields more accurate activation of relevant basis functions. In contrast, \cref{fig:nonorth} shows that non-orthogonal GP tends to greedily learn functions, such as the cubic spring term $(-k y^3)$, that are already present in the dictionary, resulting in less interpretable system identification via basis function coefficients.

\paragraph{System Identification using Partial Dictionary: }
We omit the cubic basis $y^3$ from the available dictionary to demonstrate the model's ability to learn missing structure.

\begin{equation}
\mathbf{\Phi}(\mathbf{x}) = [{y} , {\dot y} , {y}^2 , {\dot y}^2  , {\dot y}^3]^\top
\end{equation}

\begin{figure}
\includegraphics[width=\textwidth]{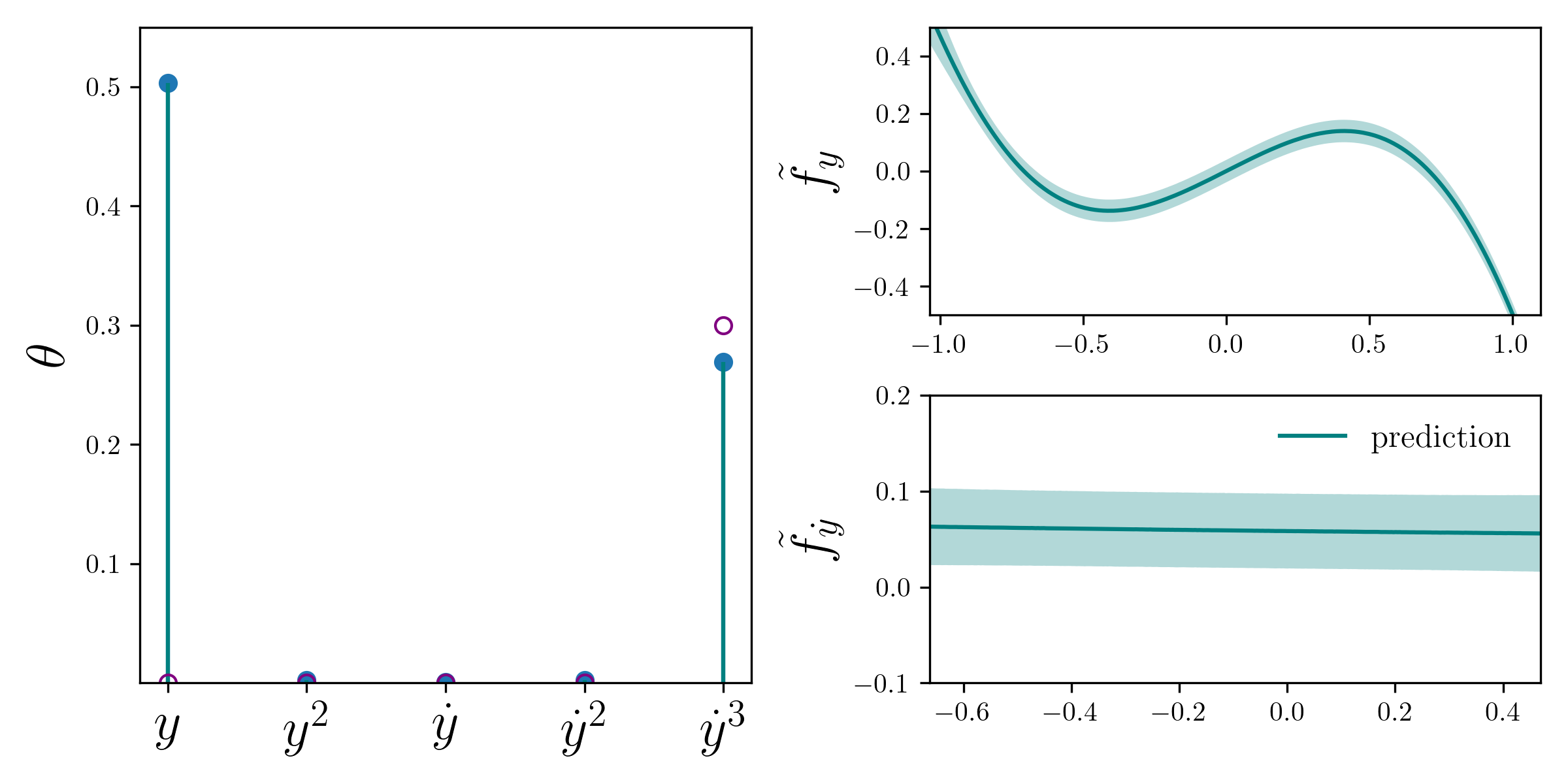}
\caption{System Identification using Orthogonal GP Grey Box Modeling for Partial Dictionary: (left panel) parameters identified via sparse regression using the white box; (right panel) the black-box discrepancy model} \label{fig:orth_cube} 

\end{figure}

\begin{figure}
\includegraphics[width=\textwidth]{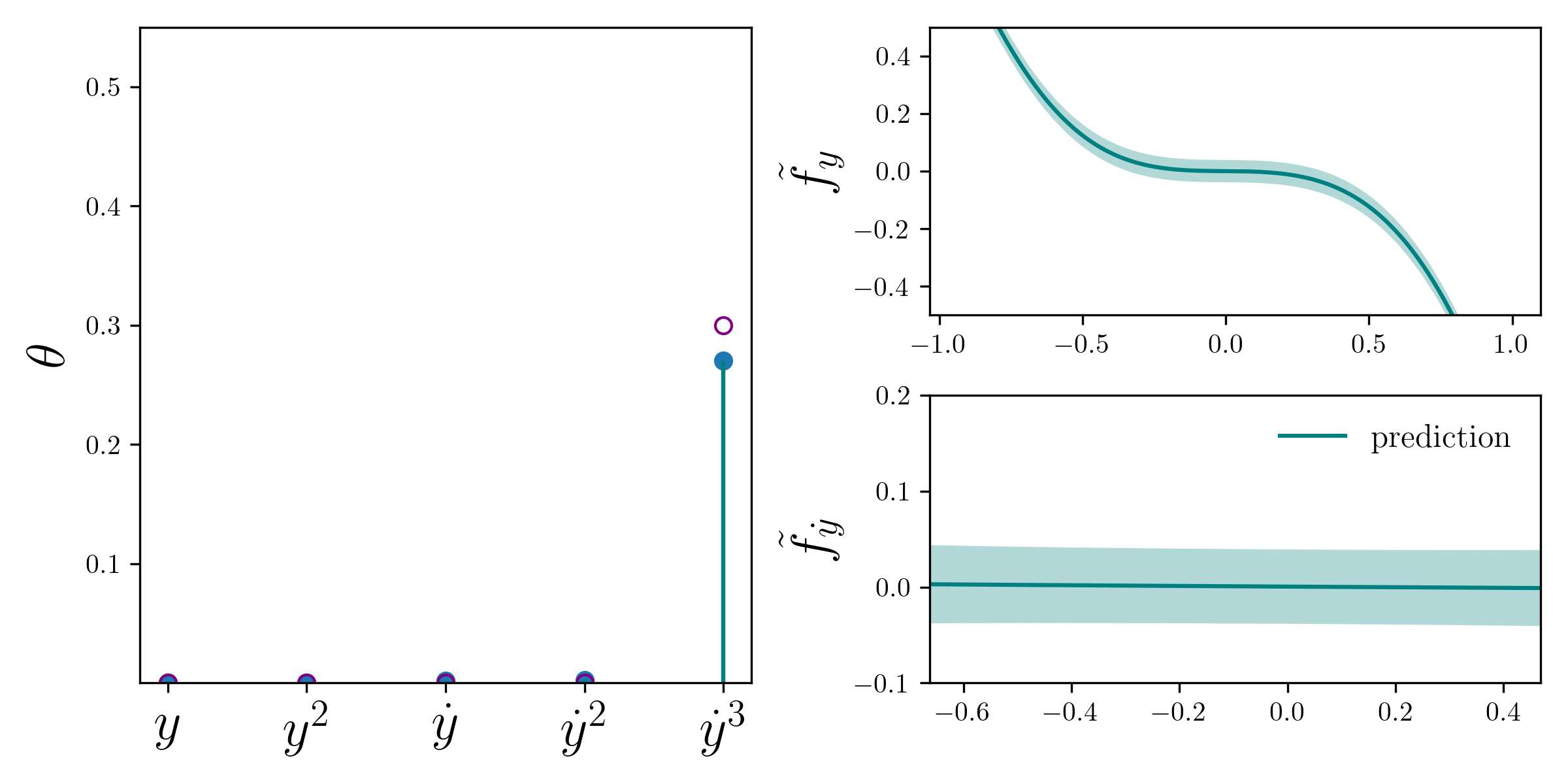}
\caption{System Identification using Non-Orthogonal GP Grey Box Modeling for Partial Dictionary: (left panel) parameters identified via sparse regression using the white box; (right panel) the black-box discrepancy model} \label{fig:nonorth_cube}
\end{figure}

\Cref{fig:orth_cube} and \Cref{fig:nonorth_cube} illustrate the results obtained using partial dictionary.  
Similar to the full dictionary case, \Cref{tab:placeholder} shows that both methods achieve comparable predictive accuracy.

From \cref{fig:nonorth_cube} and \cref{fig:orth_cube}, we observe that both Non-Orthogonal GP and Orthogonal GP identify the dictionary term $\dot{y}^3$ and utilize Gaussian Processes to model the missing $y^3$ dynamics. This allows them to accurately model the system even with incomplete dictionary, overcoming a major limitation of the standard SINDy. 

An interesting feature observed in \cref{fig:orth_cube} is that, the $y$ term also gets activated even though its not part of the true equation. This is because, in the absence of $y^3$, the $y$ term partially explains the system dynamics and hence, is chosen by the model. Since the dictionary contributes a $y$ term, the GP residual $\tilde{f}_y$ doesn't learn a perfect $y^3$. This is further proof that the orthogonal kernel prioritizes the dictionary if there is significant contribution from the basis function. 

\begin{table}
    \centering
    \caption{Table summarizes the NMSE scores of the model by evaluating the predicted acceleration against the ground-truth acceleration over the time interval 0 to 100s.}
    \begin{tabular}{lcc}
    \hline
    \textbf{Dictionary Type} & \textbf{Orthogonal GP} & \textbf{Non-Orthogonal GP} \\
    \hline
    Complete Dictionary & 
    \begin{tabular}{c}
     NMSE: $0.000039$
     \end{tabular} & 
    \begin{tabular}{c}
     NMSE: $0.001001$
     \end{tabular}
     \\
     \hline
     Partial Dictionary &
    \begin{tabular}{c}
     NMSE: $0.000014$
     \end{tabular}      &
    \begin{tabular}{c}
     NMSE: $0.000089$
     \end{tabular}      
     \\
     \hline
    \end{tabular}
    \label{tab:placeholder}
\end{table}

\section{Conclusion}

We implement an orthogonality-based model in which the structure learned by the black-box component is constrained to be orthogonal to the physics-informed dictionary. %
The method naturally balances interpretability with complexity, fitting data while identifying system parameters from the dictionary of bases. %
This initial work presents the method applied to a simulated (numerical) nonlinear oscillator.

\begin{credits}
\subsubsection{\ackname} This study was funded
by a feasibility project of the SofTMech Statistical Emulation and Translation Hub (EP/T017899/1) and the Digital Twinning NetworkPlus: DTNet+ (EP/Y016289/1).

\subsubsection{\discintname}
None
\end{credits}

\bibliographystyle{splncs04}
\bibliography{ref}

\end{document}